\documentclass[letterpaper]{article}
\usepackage{iccc}

\usepackage[utf8]{inputenc}

\usepackage{times}
\usepackage{helvet}
\usepackage{courier}

\usepackage{soul}

\usepackage{url}
\usepackage[hidelinks]{hyperref}
\usepackage{graphicx}
\usepackage[small]{caption}
\usepackage{subcaption}
\usepackage[export]{adjustbox}

\usepackage{amsmath}
\usepackage{amsfonts}
\usepackage{amsthm}
\usepackage{amssymb,mathtools}

\usepackage{booktabs}
\usepackage{algorithm}
\usepackage{algorithmic}
\usepackage{natbib}
\usepackage{flushend} 

\usepackage[nolist]{acronym}

\usepackage{multirow}

\usepackage[nameinlink]{cleveref}
\creflabelformat{equation}{#2\textup{#1}#3}  

\newcommand{\nowordbreak}[1]{\mbox{#1}}

\def\doitourl#1{http://dx.doi.org/#1}

\def\ArXivtourl#1{http://arxiv.org/abs/#1}

\newcommand{\mathbold}[1]{\ensuremath{\boldsymbol{\mathbf{#1}}}}
\newcommand{\mbp}{\mathbold{p}}
\newcommand{\mbw}{\mathbold{w}}
\newcommand{\mbx}{\mathbold{x}}
\newcommand{\mbC}{\mathbold{C}}
\newcommand{\mbK}{\mathbold{K}}
\newcommand{\mbQ}{\mathbold{Q}}

\newcommand{\mblambda}{\mathbold{\lambda}}
\newcommand{\mbLambda}{\mathbold{\Lambda}}

\newcommand{\R}{\mathbb{R}}  
\newcommand{\diag}{\textrm{diag}}
\newcommand{\T}{^{\mkern-1.5mu\mathsf{T}}}  

\newcommand{\name}{Towards Mode Balancing of Generative Models via Diversity Weights}

\title{\name}

\author{
    Sebastian Berns\,\textsuperscript{1}{\rm ,}
    Simon Colton\,\textsuperscript{1} \and 
    Christian Guckelsberger\,\textsuperscript{2,\,1}\\
    \textsuperscript{1}\,School of Electronic Engineering and Computer Science, Queen Mary University of London, UK\\
    \textsuperscript{2}\,Department of Computer Science, Aalto University, Espoo, Finland
}

\begin{document} 
\maketitle
\begin{abstract}
    \begin{quote}
        Large data-driven image models are extensively used to support creative and artistic work. Under the currently predominant distribution-fitting paradigm, a dataset is treated as ground truth to be approximated as closely as possible. Yet, many creative applications demand a diverse range of output, and creators often strive to actively diverge from a given data distribution. We argue that an adjustment of modelling objectives, from pure mode coverage towards mode balancing, is necessary to accommodate the goal of higher output diversity. 
We present \emph{diversity weights}, a training scheme that increases a model’s output diversity by balancing the modes in the training dataset. First experiments in a controlled setting demonstrate the potential of our method. 
We discuss connections of our approach to 
\acl{DEI} in generative machine learning more generally, and \acs{CC} specifically.
An implementation of our algorithm is available at \url{https://github.com/sebastianberns/diversity-weights}

    \end{quote}
\end{abstract}

\section{Introduction}

\Acp{LIGM}, in particular as part of text-to-image generation systems \citep{ramesh2021zeroshot,saharia2022photorealistic}, have been widely adopted by visual artists to support their creative work in art production, ideation, and visualisation \citep{ko2023large,vimpari2023TTIG}. 
While providing vast possibility spaces, \acp{LIGM}, trained on huge image datasets scraped from the internet, not only adopt but often exacerbate data biases, as observed in word embedding and captioning models \citep{bolukbasi2016man,zhao2017men,hendricks2018women}. The tendency to emphasise majority features and to primarily reproduce the predominant types of data examples can be limiting for many \ac{CC} applications that use machine learning-based generators \citep{loughran2017application}. Learned models are often used to illuminate a possibility space and to produce artefacts for further design iterations. Examples range from artistic creativity, like the production of video game assets \citep{liapis2014computational,volz2018evolving}, over constrained creativity, e.g.~industrial design and architecture \citep{bradner2014parameters}, to scientific creativity, such as drug discovery \citep{madani2023large}.
Many of these and similar applications would benefit from higher diversity in model output. Given that novelty, which underlies diversity, is considered one of the essential aspects of creativity \citep{boden2004creative,runco2012standard}, we expect that, vice versa, a stronger focus on diversity can also foster creativity \citep[cf.][]{stanley2015greatness}.  

Most common modelling techniques, however, follow a distribution-fitting paradigm and do not accommodate the goal of higher diversity. Within this paradigm, one of the primary generative modelling objectives is \emph{mode coverage}~\citep{zhong2019rethinking}, i.e.~the capability of a model to generate all prominent types of examples present in a dataset. While such a model can in principle produce many types of artefacts, it does not do so reliably or evenly.
A model’s probability mass is assigned in accordance to the prevalence of a type of example or feature in a dataset. Common examples or features 
have higher likelihood under the model than rare ones. 
As a consequence, samples with minority features are not only less likely to be obtained by randomly sampling a model, they are also of lower fidelity, e.g.~in terms of image quality.
Related studies on Transformer-based language models \citep{razeghi2022impact,kandpal2022deduplicating} have identified a “superlinear” relationship: while training examples with multiple duplicates are generated “dramatically more frequently”, examples that only appear once in the dataset are rarely reproduced by the model.

In this work, we argue for an adjustment of modelling techniques from mode coverage to \emph{mode balancing} to enrich \ac{CC} with higher output diversity. Our approach allows to train models that cover all types of training examples and can generate them with even probability and fidelity.
We present a two-step training scheme designed to reliably increase output diversity. 
Our technical contributions are:
\begin{itemize}
    \item \emph{Diversity weights}, a training scheme to increase a generative model’s output diversity by taking into account the relative contribution of individual training examples to overall diversity.  
    \item \emph{\acf{wFID}}, an adaptation of the \acs{FID} measure to estimate the distance between a model distribution and a target distribution modified by weights over individual training examples.
    \item A proof-of-concept study, demonstrating the capacity of our method to increase diversity, examining the trade-off between artefact typicality and diversity.
\end{itemize}

\noindent
In the following sections, we first introduce the objective of \emph{mode balancing} and highlight its importance for \ac{CC} based on existing frameworks and theories. 
Then, we provide background information on the techniques relevant for our work.
Next, we present our \emph{diversity weights} method in detail, as well as our formulation of \emph{Weighted \acs{FID}}.
Following this, we present the setup and methodology of our study and evaluate its results.
In the discussion section, we contribute to the debate on issues of \ac{DEI} in generative machine learning more generally, and \ac{CC} specifically, by explaining how our method could be beneficial in addressing data imbalance bias.
This is followed by an overview of related work, our conclusions and an outlook on future work.

\section{Mode Balancing}

\begin{figure}[t]
    \centering
    \includegraphics[width=1.0\linewidth]{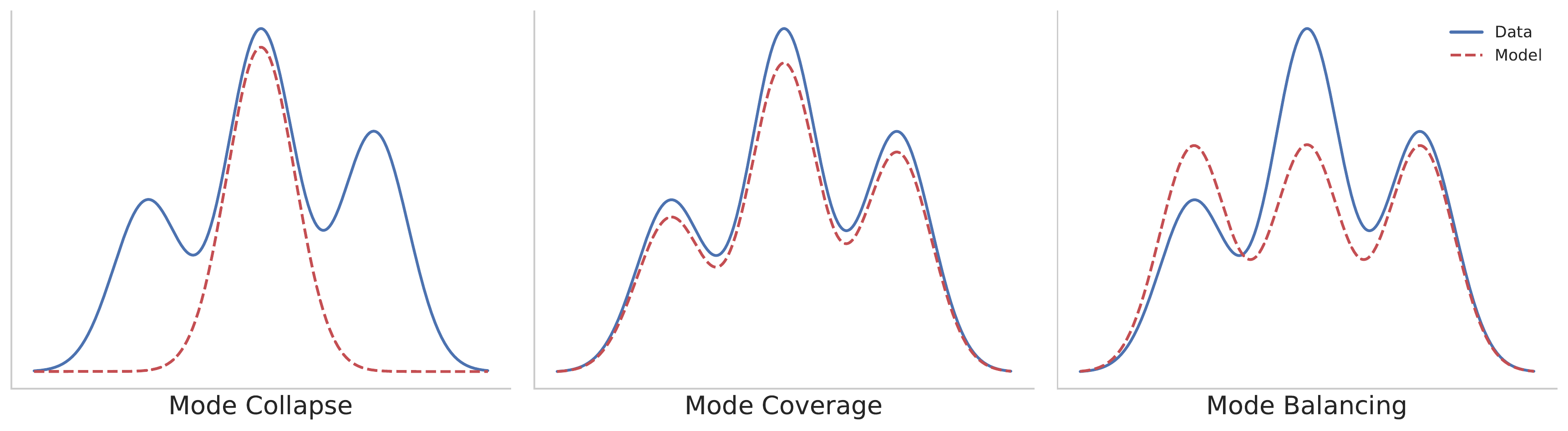}
    \caption{\emph{Mode collapse}: the model does not cover all modes in the data distribution. \emph{Mode coverage}: the data distribution’s modes are modelled as closely as possible w.r.t. their likelihood. \emph{Mode balancing}: the model covers all modes, but with equal likelihood.}
    \label{fig:modebalancing}
\end{figure}

Generative deep learning models now form an integral part of \ac{CC} systems \citep{berns2021framework}.
A lot of work on such models is concerned with \emph{mode coverage}: to match a data distribution as closely as possible by accurately modelling all types of examples in a dataset (\cref{fig:modebalancing}). In the specific case of \acp{GAN}, great effort is put into preventing \emph{mode collapse}, a training failure state in which a model disregards important modes and is only able to produce a few types of training examples. Mode coverage is captured formally in common evaluation measures such as \ac{FID} and \ac{PR}. Crucially, this is always done in reference to the training set statistics or data manifold.
In this context, diversity is often arguably misused to refer to mode coverage. While mode coverage describes the fraction of modes in a dataset that are represented by a model, the diversity of a model’s output, if understood more generally and intuitively, can theoretically be higher than that of the dataset.

Mode coverage is conceptually similar to the notion of \emph{typicality}~\citep{ritchie2007empirical}. Defined as the extent to which a produced output is “an example of the artefact class in question”, a model which only generates outputs with high typicality, if sampled at random, has to provide most support to those training set examples with the highest density of features characteristic of that artefact, i.e.~to maximise mode coverage. Crucially, sampling from the model would resemble going along the most well-trodden paths in the possibility space defined by the dataset and, as Ritchie already suggests, counteract novelty as a core component of creativity \citep{boden2004creative,runco2012standard}.

Crucially, mode balancing breaks with the convention of viewing the dataset as ‘ground truth’. Instead, we consider the dataset to provide useful domain information and the characteristics of \emph{typical} examples~\citep{ritchie2007empirical}. 
But a data distribution does not have to be matched exactly. Particularly in artistic applications, creators often strive to \emph{actively diverge} from the typical examples in a dataset \citep{berns2020bridging,broad2021active}. 
To stay with our metaphor, borrowed from \citet{veale2019systematizing}, \emph{mode balancing} allows us to walk more along the less trodden paths and thus especially support exploratory and transformational creativity \citep{boden2004creative,stanley2015greatness}. 
In contrast to the mode coverage paradigm, in mode balancing, diversity is measured independently of the training data distribution. In the theoretical case of a balanced dataset of absolutely dissimilar examples, i.e.~multiple equally likely modes, our method would assign uniform weights to all examples and thus be identical to standard training schemes with random sampling.

\section{Background}

\subsection{Probability-Weighted Vendi Score}

We adopt the \acf{VS} as a measure of dataset diversity and employ its probability-weighted formulation in our work \citep{friedman2022vendi}.
Given a set of artefacts $x_1, \dots, x_n$, the probability-weighted \ac{VS} is based on a probability vector $\mbp=(p_1, \dots, p_n)$ and a similarity matrix $\mbK\in\R^{\,n\times n}$ between pairs of artefacts such that $\mbK_{ii} = 1$.
Calculating \ac{VS} involves various steps.
First, the probability-weighted similarity matrix is defined as $\mbK^{\mbp}=\diag(\sqrt{\mbp})\,\mbK\,\diag(\sqrt{\mbp})$. Its eigenvectors $\lambda_1, \dots, \lambda_n$ can be obtained via the eigendecomposition $\mbK^{\mbp} = \mbQ \mbLambda \mbQ^{-1}$, where $\mblambda = \diag(\mbLambda)$.
The probability-weighted \acf{VS} is the exponential of the Shannon entropy of the eigenvalues of the probability-weighted similarity matrix:
\begin{equation}
    \label{eq:prob-vendi}
    \textrm{VS}(\mbK, \mbp) = \exp \Big( -\sum^n \lambda_i \, \log \lambda_i \Big)
\end{equation}
Also known as \emph{perplexity}, exponential entropy can be used to measure how well a probability model predicts a sample. Low perplexity indicates good prediction performance. Consequently, the more diverse a sample, the more difficult its prediction, the higher the perplexity and its \ac{VS}.

\subsubsection{Illustrative Example}
The probability vector $p$ represents the relative abundances of individual artefacts. Instead of repeating identical artefacts in a set, their prevalence can be expressed with higher probability. For illustration, we present an example of four artefacts, of which three are absolutely similar to each other and one is absolutely dissimilar to all others. All have equal probability.
\begin{equation}
    \label{eq:example-a}
    \mbK^a = \begin{psmallmatrix}
        1 & 0 & 0 & 0 \\
        0 & 1 & 1 & 1 \\
        0 & 1 & 1 & 1 \\
        0 & 1 & 1 & 1 \\
    \end{psmallmatrix},\quad
    \mbp^a = \begin{psmallmatrix}
        0.25 \\
        0.25 \\
        0.25 \\
        0.25 \\
    \end{psmallmatrix}
\end{equation}
The same information can be reduced to two absolutely dissimilar artefacts and the corresponding probabilities $\mbp^b$.
\begin{equation}
    \label{eq:example-b}
    \mbK^b = \begin{psmallmatrix}
        1 & 0 \\
        0 & 1 \\
    \end{psmallmatrix},\quad
    \mbp^b = \begin{psmallmatrix}
        0.25 \\
        0.75 \\
    \end{psmallmatrix}
\end{equation}
Both representations yield the same \ac{VS}, which reflects the imbalanced set of two absolutely dissimilar artefacts.
$\textrm{VS}(\mbK^a, \mbp^a) = \textrm{VS}(\mbK^b, \mbp^b) = 1.755\dots$

The imbalance of our example set negatively affects its diversity. If all items in the set are given equal importance, one artefact is under-represented. Instead, each of the two absolutely dissimilar artefacts in the set should thus be assigned equal weight $p=0.5$. In the case of repetitions, this weight has to be divided across the repeated artefacts.
\begin{equation}
    \label{eq:example-c}
    \mbK^c = \begin{psmallmatrix}
        1 & 0 & 0 & 0 \\
        0 & 1 & 1 & 1 \\
        0 & 1 & 1 & 1 \\
        0 & 1 & 1 & 1 \\
    \end{psmallmatrix},\quad
    \mbp^c = \begin{psmallmatrix}
        0.5 \\
        0.166\dots \\
        0.166\dots \\
        0.166\dots \\
    \end{psmallmatrix}
\end{equation}
This maximises \ac{VS} to reflect the effective number of absolutely dissimilar artefacts $\textrm{VS}(\mbK^c, \mbp^c) = 2$.

\subsection{Importance Sampling}

Conventionally, training examples are drawn from a dataset with uniform probability. In importance sampling, instead, examples are chosen according to their contribution to an unknown target distribution. 
In our case, the importance of training examples is determined by their individual contribution to the overall dataset diversity as quantified by the optimised probability distribution $p$ (see example above).
We aim to increase the output diversity of a model. For this, we replace the basic sampling operation by a diversity-weighted importance sampling scheme.

\subsection{Model Evaluation}

To assess model performance, we use some common measures for generative models, as well as measures specifically relevant to our method.
\acf{IS} \citep{salimans2016improved}, \acf{FID} \citep{heusel2017gans}, and Precision-Recall (\nowordbreak{k-NN} parameter \nowordbreak{$k=3$})~\citep{kynkaanniemi2019improved} quantify sample fidelity and mode coverage w.r.t. the unbiased training data distribution. 
We employ our \acf{wFID} to account for the change in target distribution, induced by our method through diversity-weighted sampling (see below for details).
Diversity is estimated with the \acf{VS} \citep{friedman2022vendi}.

Note that we follow the recommendations by \citet{barratt2018note} and calculate \ac{IS} over the entire generated set of samples, removing the common split into subsets. We also remove the exponential, such that the score becomes interpretable in terms of mutual information.
While not all reported scores are directly comparable to other works, our measurements are internally consistent and reliable.

\subsubsection{Image Embeddings}

Instead of comparing image data on raw pixels, standard evaluation measures of model performance have relied on image classification networks to be used as embedding models for feature extraction. The InceptionV3 model \citep{szegedy2016rethinking} is most commonly used as a representative feature space and has been widely adopted as part of a standard measurement pipeline.
Unfortunately, small numerical differences in model weights, implementations and interpolation operations can compound to bigger discrepancies. For example, image scaling to match the input size of an embedding model can change the computed features and thus affect the subsequent measurements \citep{parmar2022aliased}. Furthermore, embedding models trained on the ImageNet dataset, like InceptionV3, inherit the dataset’s biases, which can lead to unreliable measurements that do not agree with human assessment \citep{kynkaanniemi2023role}.
In this work, we therefore follow the recommendations for anti-aliasing re-scaling and use \acs{CLIP}~ViT-L/14~\citep{radford2021learninga} as the image embedding model in our feature extraction and measurement pipelines (except for \ac{IS}).
Note that, while trained on a much larger (proprietary) dataset and better suited as embedding model, \acs{CLIP} still has its own biases.

\section{Diversity Weights}

If artefacts in a set are repeated, i.e.~their relative abundance is increased, their individual contribution to the overall diversity of the set decreases. Yet, with uniform weighting, all artefacts contribute to the model distribution equally (cf.~\cref{eq:example-a}). Instead, we aim to adjust the weight of individual artefacts in a set in accordance with their contribution to overall diversity. 

We formulate an optimisation problem to find the optimal weight for each artefact in a set, such that its diversity, as measured by \ac{VS}, is maximised.
\begin{align}
    & \max \, \exp \Big( - \sum^n \lambda_i \log \lambda_i \Big) \\
    \text{s.t. } & \sum^n p_i = 1 \qquad 0 \leq p_i \leq 1 \nonumber\\
    \text{where}\enspace & \mbp = (p_i, \dots, p_n),\enspace p_i \in \R^{[0, 1]} \nonumber\\
                  & \mbK\in\R^{\,n\times n},\enspace \mbK_{ii} = 1 \nonumber\\
                  & \mbK^{\mbp} = \diag(\sqrt{\mbp})\,\mbK\,\diag(\sqrt{\mbp})
                  = \mbQ\mbLambda\mbQ^{-1} \nonumber\\
                  & \mblambda = \diag(\mbLambda) = (\lambda_1, \dots, \lambda_n) \nonumber
\end{align}
\vspace{0.1em}

\subsection[Optimisation Algorithm]{Optimisation Algorithm\footnote{An implementation of the optimisation algorithm is available at \url{https://github.com/sebastianberns/diversity-weights}}}

We compute an approximate solution to the optimisation problem via gradient descent (\cref{alg:diversityweights}). 
The objective function consists of two terms: diversity loss and entropy loss. The diversity loss is defined as the negative probability-weighted \ac{VS} of the set of artefacts, given its similarity matrix and the corresponding probability vector (cf.~\cref{eq:prob-vendi}). 
To ensure the optimised artefact probability distribution follows the \citet{kolmogoroff1933grundbegriffe} axioms, we make the following adjustments.
Instead of optimising the artefact probabilities directly, we optimise a weight vector $\mbw$. The probability vector $\mbp$ is obtained by dividing the $\mbw$ by the sum of its values, which guarantees the second axiom.
To satisfy the first axiom, we implement a fully differentiable version of \ac{VS} in log space. Optimising in log space enforces weights above zero, since the logarithm $\log x$ is only defined for $x > 0$ and tends to negative infinity as $x$ approaches zero. 
However, if the weights have no upper limit, values can grow unbounded. A heavy-tailed weight distribution negatively affects the importance sampling step of our method during training, as batches can become saturated with the highest-weighted training examples, causing overfitting. We therefore add an entropy loss term $\mathrm{H}(\mbp)=-\sum p_i \log(p_i)$ to be maximised in conjunction with the diversity loss. The entropy loss acts as a regularisation term over the weight vector, such that its distribution is kept as close to uniform as possible.
The emphasis on the two loss terms is balanced by the hyperparameter $\gamma\in [0, 1]$.
\begin{equation}
    \mathcal{L} = -\,\gamma\mathrm{VS}(\mbK, \mbp) - (\gamma-1)\,\mathrm{H}(\mbp)
    \text{, }\quad\enspace
    \mbp = \frac{\mbw}{||\mbw||_1}
\end{equation}
Given a normalised data matrix $X$ where rows are examples and columns are features, we obtain the similarity matrix $\mbK$ by computing the Gram matrix $K = X\cdot X\T$. The weight vector $\mbw$ is initialised with uniform weights $w_i=\log(1)=0$. The probability vector $\mbp$ is obtained by dividing the weight vector $\mbw$ by the sum of its values.
We choose the Adam optimiser \citep{kingma2015adam} with $\beta_1=0.9$ and $\beta_2=0.999$. The learning rate decays exponentially every $5$ iterations by a factor of $0.99$.

\begin{algorithm}[t]
    \caption{Vendi Score Diversity Weight Optimisation}
    \label{alg:diversityweights}
    \textbf{Input}: Similarity matrix $\mbK$ of $N$ artefacts\\
    \textbf{Parameter}: 
        Loss term balance~$\gamma$, 
        num iterations~$I$, 
        learning rate~$\alpha$, 
        Adam hyperparams~$\beta_1$,~$\beta_2$
    \begin{algorithmic}[1] 
        \STATE Initialise $\mbw = (w_1, \dots, w_N)$, where $w_i=1$
        \FOR{$i=0$ \TO $I$}
            \STATE $\mbp \gets \mbw / \sum w_i$
            \STATE $g \gets -\nabla_{\mbp}\, \gamma \mathrm{VS}(\mbK, \mbp) - (\gamma-1)\, \mathrm{H}(\mbp)$
            \STATE $\mbw \gets \mathrm{Adam}(\mbw, g, \alpha, \beta_1, \beta_2) $
        \ENDFOR
    \end{algorithmic}
    \textbf{Output}: Weight vector $\mbw$
\end{algorithm}

\subsection{Weighted FID}

The performance of generative models, in particular that of implicit models like GANs, is conventionally evaluated with the \ac{FID} \citep{heusel2017gans}. Raw pixel images are embedded into a representation space, typically of an artificial neural network. Assuming multi-variate normality of the embeddings, \ac{FID} then estimates the distance between the model distribution and the data distributions from their sample means and covariance matrices.

In our proposed method, however, the learned distribution is modelled on a weighted version of the dataset. Moreover, referring to the standard statistics of the original dataset is no longer applicable, as the weighted sampling scheme changes the target distribution.
We therefore adjust the measure such that it becomes the \acf{wFID}, where the standard mean and covariances to calculate the dataset statistics are substituted by the weighted mean $\mu^* = \big( \sum w_i \mbx_i \big) / \sum w_i$ and the weighted sample covariance $\mbC = \big( \sum w_i (\mbx_i - \mu^*)\T (\mbx_i - \mu^*) \big) / \sum w_i$.
Note that the statistics of the model distribution need to be calculated without weights as the model should have learned the diversity-weighted target distribution.

\section{Proof-Of-Concept Study on \texorpdfstring{\\}{}Hand-Written Digits}

We show the effect of the proposed method in an illustrative study on pairs of handwritten digits. While artistically not particularly challenging, digit pairs have several benefits over other exemplary datasets.
First, the pairings of digits create a controlled setting with two known types of artefacts.
Second, hand-written digits present a simple modelling task, in which the quality and diversity of a model’s output is easy to visually assess. 
And third, generating digits is fairly uncontroversial. While, for example, generating human faces is more relevant for the subject of diversity, it is also a highly complex and potentially emotive domain.

\subsection{Methodology}

For individual pairs of digits, we quantitatively and qualitatively evaluate the results of \ac{GAN} training with diversity weights and compare it against standard training. Experiments are repeated five times with different random seeds.

\paragraph{Digit Pairs}
From the ten classes of the MNIST training set, we select three digit pairs: 0-1, 3-8, and 4-9, which represent examples of similar and dissimilar pairings. For example, images of hand-written zeros and ones are easy to distinguish, as they are either written as circles or straight lines. In contrast, threes and eights are both composed of similar circular elements.

\paragraph{Balanced Datasets}

For each pair of digits, we create five balanced datasets (with different random seeds) of 6,000 samples each. Each dataset consists of 3,000 samples of either digit, randomly selected from the MNIST training set.
We compute features by embedding all images using the \acs{CLIP} ViT-L/14 model. To optimise the corresponding diversity weights, we obtain pairwise similarities between images by calculating the Gram matrix of features.

\paragraph{Diversity Weights}

For each dataset (5 random draws per digit pair), we optimise the diversity weights for 100 iterations.
We fine-tune the loss term balance hyperparameter and determine its optimal value $\gamma=0.8$, where the weights converge to a stable distribution, while reaching a diversity loss as close to the maximum as possible. Without the entropy loss term ($\gamma=1.0$) the weights yield the highest \ac{VS}, but reach both very high and very low values. Large differences in weight values negatively affect the importance sampling step of our method during training, as batches can become saturated with the highest-weighted training examples. In contrast, a bigger emphasis on the entropy loss ($\gamma=0.6$) results in the weights distribution being closer to uniform, but does not maximise diversity. The hyperparameter $\gamma$ provides control over the trade-off between diversity and \emph{typicality}, i.e.~the extent to which an generated artefact is a typical training example~\citep{ritchie2007empirical}.
The \ac{VS} of the digit datasets when measured without and with diversity weights at different loss term balances are presented in \cref{table:vs-uniform-divw}.

\begin{figure}[t]
    \centering
    \includegraphics[width=1.0\linewidth]{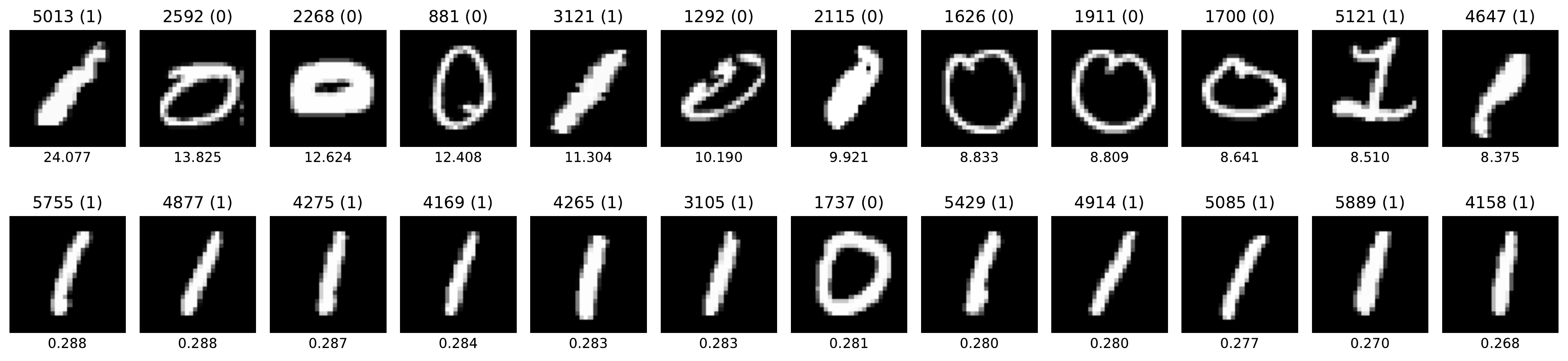}
    \includegraphics[width=1.0\linewidth]{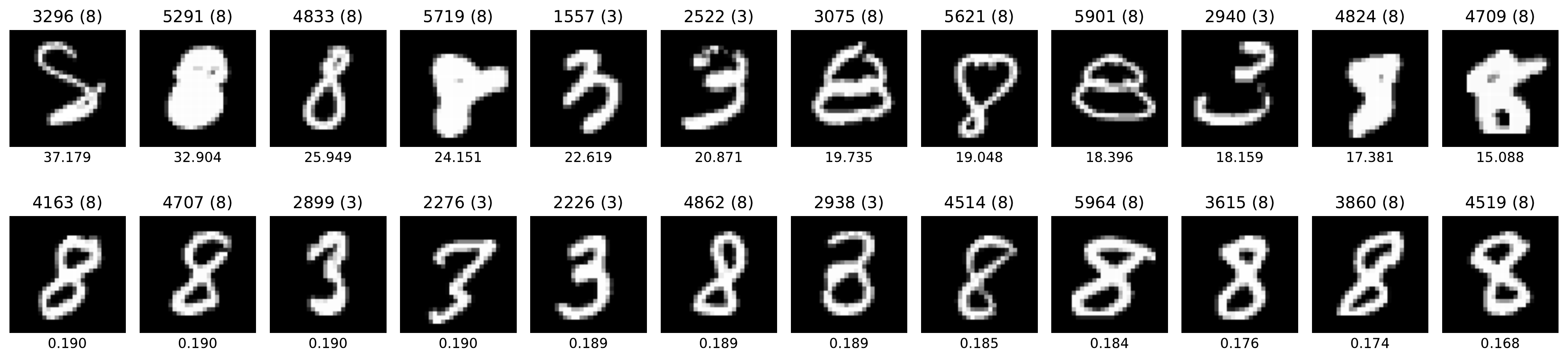}
    \includegraphics[width=1.0\linewidth]{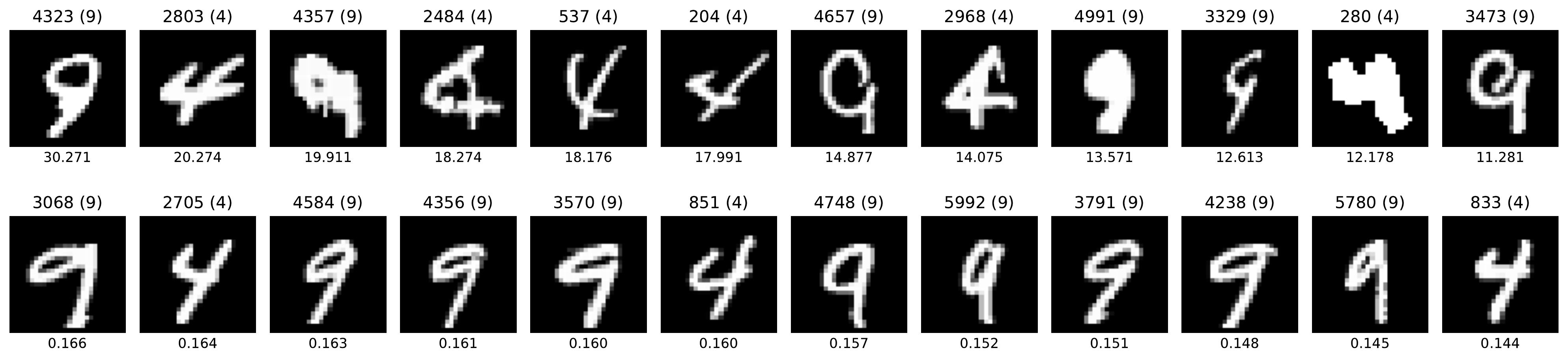}
    \caption{Digits ordered by diversity weight (index above with label in brackets, weight below). First two rows: pair 0-1, two middle rows: pair 3-8, last two rows: pair 4-9. Odd rows: twelve highest weighted, even row: twelve lowest weighted.}
    \label{fig:digitsorderedbyw}
\end{figure}

The resulting diversity weight for each of the 6,000 samples corresponds to their individual contributions to the overall diversity of the dataset. 
We give an overview of the highest and lowest weighted data samples in \cref{fig:digitsorderedbyw}. Low-weighted samples are typical examples of the MNIST dataset: e.g.~round zeros and simple straight ones, all of similar line width. High-weighted samples show a much greater diversity: thin and thick lines, imperfect circles as zeros, ones with nose and foot line.

\begin{table}[t]

\centering
\caption{\acf{VS} of digit pair datasets (mean ± std dev) with uniform and diversity weights with different loss balances $\gamma$}

\label{table:vs-uniform-divw}

\begin{tabular}{c|ccc}
\toprule
\multirow{2.5}{*}{\ac{VS} weights} &\multicolumn{3}{c}{MNIST digit pairs}\\
\cmidrule{2-4}
& Pair 0-1 & Pair 3-8 & Pair 4-9 \\
\midrule
Uniform weights
    & 1.77\scriptsize{$\pm$0.003} 
    & 1.96\scriptsize{$\pm$0.004}
    & 2.07\scriptsize{$\pm$0.004}\\

DivW ($\gamma=0.6$)
    & 2.13\scriptsize{$\pm$0.020} 
    & 2.64\scriptsize{$\pm$0.016} 
    & 2.65\scriptsize{$\pm$0.010}\\

DivW ($\gamma=0.8$)
    & 2.79\scriptsize{$\pm$0.052} 
    & 3.45\scriptsize{$\pm$0.027} 
    & 3.38\scriptsize{$\pm$0.025}\\
    
DivW ($\gamma=1.0$)
    & 3.08\scriptsize{$\pm$0.046} 
    & 3.67\scriptsize{$\pm$0.023} 
    & 3.60\scriptsize{$\pm$0.023}\\
\bottomrule
\end{tabular}

\end{table}

\paragraph{Training}
For each digit dataset, we compare two training schemes: 1) a baseline model with the standard training scheme, and 2) three models trained with our \acf{DivW} method and different loss term balances ($\gamma$), where training examples are drawn according to the corresponding diversity weights. The compared loss term balances are $\gamma=0.6$,  $\gamma=0.8$, and  $\gamma=1.0$.
All models have identical architectures \citep[\acl{WGAN-GP};][]{gulrajani2017improved} and hyperparameters and are optimised for 6,000 steps (see appendix for details).

To allow our method to develop its full potential, we increase the batch size to 6,000 samples, the size of the dataset. Training examples are drawn according to diversity weights \emph{with} replacement, i.e.~the same example can be included in a batch more than once. Small batches in turn would be dominated by the highest-weighted examples, causing overfitting and ultimately mode collapse.

\paragraph{Evaluation}
We evaluate individual models on six measures: \acf{VS} to quantify output diversity; \acf{IS}, \acf{FID} and weighted \acs{FID} (\acs{wFID}), as well as \acf{PR} to estimate sample fidelity and mode coverage. From each model we obtain 6,000 random samples, the same amount as a digit dataset. As described above, for all measures, except \ac{IS}, we use \acs{CLIP} as the image embedding model to compute image features. For \ac{VS}, we obtain pairwise similarities between images by calculating the Gram matrix of features. Our proposed \acs{wFID} measure accounts for the different target distribution induced by the diversity weights.

\subsection{Results}

\begin{figure*}[t]
    \centering
    \includegraphics[width=1.0\linewidth]{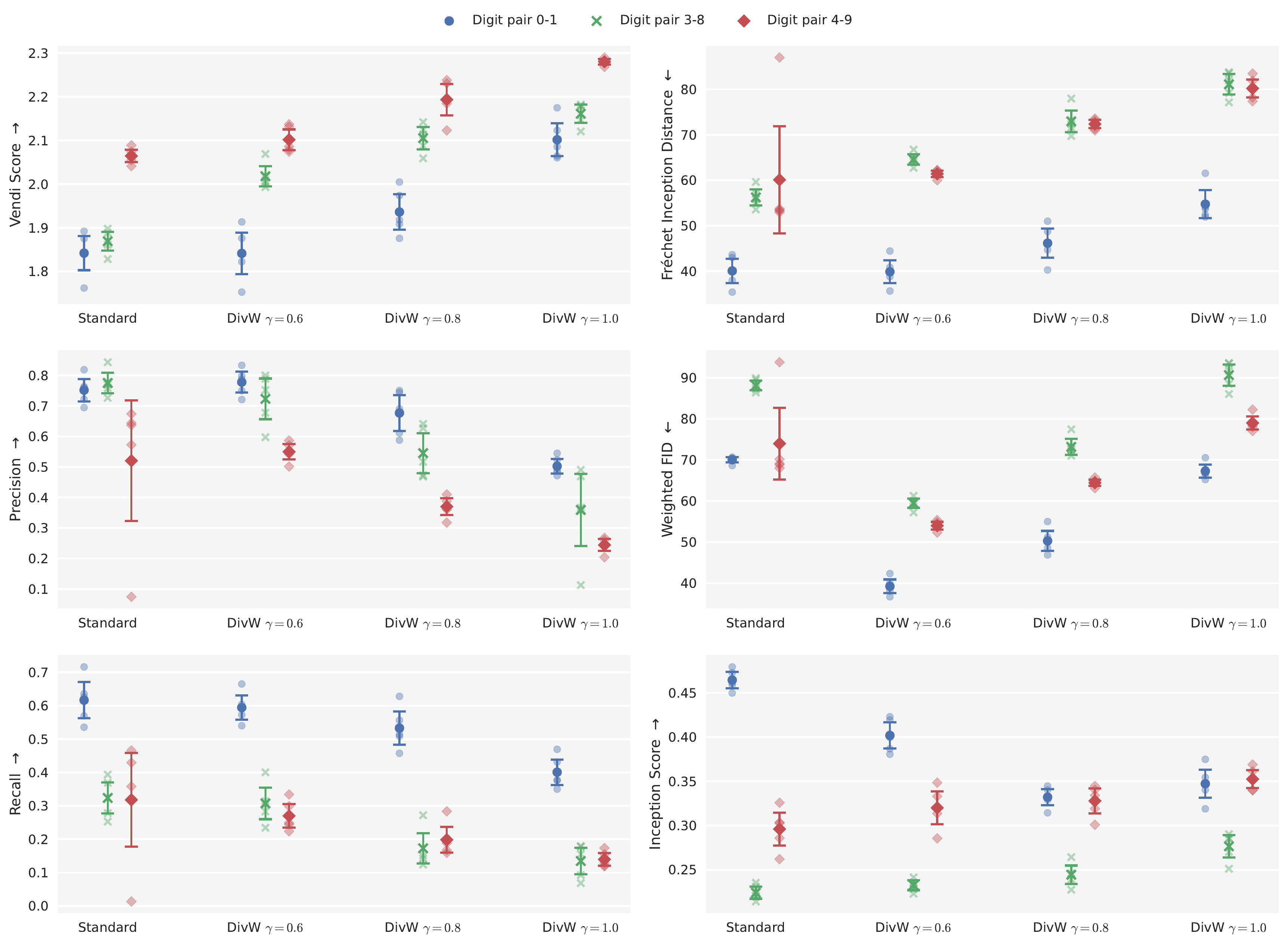}
    \caption{Performance comparison of our method (\acs{DivW}) with different loss term balances ($\gamma$) against a standard \acs{GAN}, trained on three digit pair datasets (blue circles:~0-1, green crosses:~3-8, red diamonds:~4-9) with six measures: \ac{VS}, \ac{PR} and \ac{IS} (higher is better), as well as standard FID and weighted FID scores (lower is better). Means and 95{\kern 0.1em}\% confidence intervals over five random seeds. Individual datapoints show means over five random sampling repetitions. The hyperparameter $\gamma$ provides control over the trade-off between diversity and typicality.}
    \label{fig:results-measures}
\end{figure*}

An overview of our quantitative results in given in \cref{fig:results-measures}. For three pairs of digits, we compare our \acf{DivW} method with three different loss term balances ($\gamma$) against a standard \ac{GAN}. The balance of loss terms determines the emphasis on a uniform distribution of weights (lower $\gamma$) over higher diversity (higher $\gamma$). Accordingly, in the diversity weight optimisation, a balance of $\gamma=1.0$ corresponds to a full emphasis on diversity and no entropy loss, while $\gamma=0.5$ strikes an equal balance between the two.

Our results agree on almost all measures across all three digit pairs, except on \ac{IS} which we discuss further below.
As expected, the higher the emphasis on the diversity loss, the higher (and better) the \ac{VS} (\cref{fig:results-measures}, top left). This comes with a trade-off in sample fidelity and mode coverage, as quantified by \ac{PR} (\cref{fig:results-measures}, middle and bottom left) and \ac{FID} (\cref{fig:results-measures}, top right). However, when accounting for a weighted training dataset with our Weighted \acs{FID} measure, the distance of our \acs{DivW} model distribution to the target distribution is notably lower than or at least on par with the standard model (\cref{fig:results-measures}, middle right).

Results on \ac{IS} (\cref{fig:results-measures}, bottom right) show the difficulty in distinguishing different pairs of digits. For the pairing 0-1, the standard model and the \acs{DivW} $\gamma=0.6$ model score notably higher than the other two \acs{DivW} models ($\gamma=0.8$ and $\gamma=1.0$), while their scores are lower for the pairings 3-8 and 4-9. This suggests that, even for the standard model it is difficult to model two similar digits like 3-8 and 4-9.

\begin{figure*}[t!]
\centering

\begin{tabular}{lcccc}

& 
{Standard} & 
{\acs{DivW} $\gamma=0.6$} & 
{\acs{DivW} $\gamma=0.8$} & 
{\acs{DivW} $\gamma=1.0$} 
\vspace{.5em} \\

{Digits 0-1} & 
\includegraphics[width=.2\linewidth,valign=m]{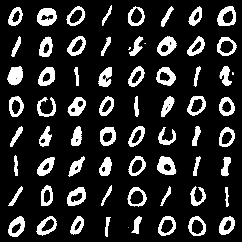} & 
\includegraphics[width=.2\linewidth,valign=m]{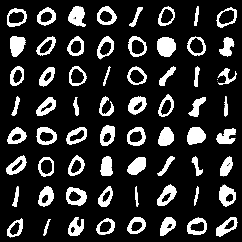} & 
\includegraphics[width=.2\linewidth,valign=m]{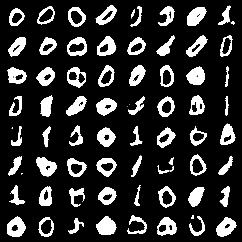} & 
\includegraphics[width=.2\linewidth,valign=m]{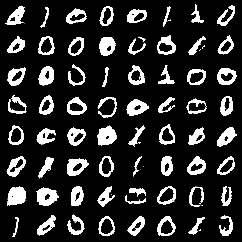} \\ \\

{Digits 3-8} & 
\includegraphics[width=.2\linewidth,valign=m]{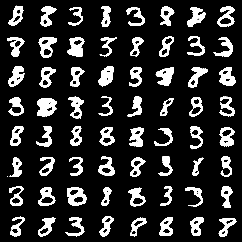} & 
\includegraphics[width=.2\linewidth,valign=m]{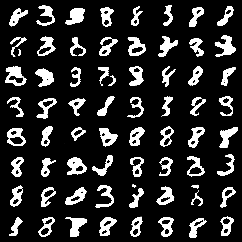} & 
\includegraphics[width=.2\linewidth,valign=m]{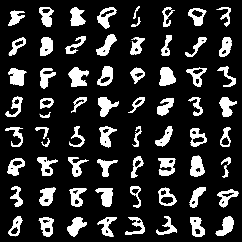} & 
\includegraphics[width=.2\linewidth,valign=m]{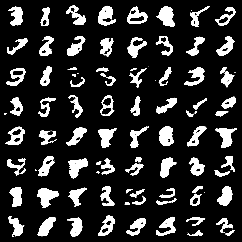} \\ \\

{Digits 4-9} & 
\includegraphics[width=.2\linewidth,valign=m]{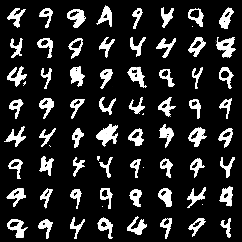} & 
\includegraphics[width=.2\linewidth,valign=m]{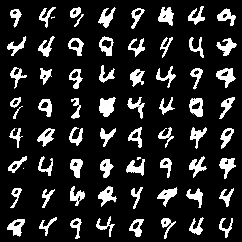} & 
\includegraphics[width=.2\linewidth,valign=m]{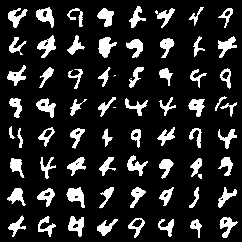} & 
\includegraphics[width=.2\linewidth,valign=m]{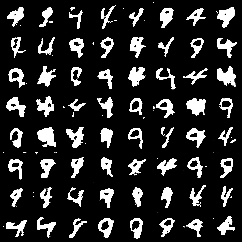}

\end{tabular}

\caption{Random samples for all digit pairs (top row: 0-1, middle: 3-8, bottom: 4-9) from the standard models (left column) and our \acs{DivW} models with different loss balances ($\gamma$). The hyperparameter $\gamma$ provides control over the trade-off between diversity and typicality.}
\label{fig:results-digits}
\end{figure*}

For visual inspection and qualitative analysis, we provide random samples in \cref{fig:results-digits} for all digit pairs and models.

\section{Discussion}

In recent years, research communities have become better aware of data biases and their impact on society through the proliferation of data-driven technologies. Likewise, \ac{CC} researchers have highlighted its potential implications for \ac{CC} research and the importance of mitigation \citep{smith2017computational,loughran2022bias}.
Real-world datasets are limited sample from a complex world and should not be considered as the ‘ground truth’, or as representing the ‘true’ distribution. This practical impossibility further motivates our proposal to shift away from the predominant mode coverage paradigm.

Ongoing debates have not yet resulted in a uniformly accepted way of dealing with data bias in generative machine learning more generally, and \ac{CC} specifically. 
One way to address data bias is to gather more or better data. But this is not always possible or practical, since collecting, curating and pre-processing new data is notoriously laborious, costly, or subject to limited access.
Another way is to instead adjust the methodology of learning from data, such that a known data bias is mitigated.
In this work, we focus on the latter and propose the \emph{diversity-weighted sampling} scheme to address the imbalance of representation between majority and minority features in a dataset. 

Diversity weights address the specific bias of \emph{data imbalance}, particularly in unsupervised learning. In contrast to supervised settings, where class labels provide a clear categorisation of training examples, here common features are often shared between various types of examples. 
This makes it difficult to find an appropriate balance of training examples. Diversity weights give an indication of which type of examples are under-represented from a diversity-maximisation perspective.
We draw a connection to issues of \ac{DEI} as data biases often negatively affect under-represented groups \citep{bolukbasi2016man,zhao2017men,hendricks2018women,stock2018convnets}.

Combining image generation models with multi-modal embedding models, like \acs{CLIP}, enables complex text-to-image generative systems which can be doubly affected by data bias through the use of two data-driven models: the image generator and the image-text embedding.
The discussion on embedding models, and other methods that can guide the search for artefacts, is beyond the scope of this paper. 
Our work focuses on the image generators powering these technologies. Yet, a conscious shift to \emph{mode balancing}, in particular for the training of the underlying generative model, could support the mitigation of bias in text-to-image generation models, complementing existing efforts in prompt engineering after training \citep{colton:nesy22}.

It is worth noting, that our method also introduces bias, particularly emphasising under-represented features in the dataset. We do this explicitly and for a specific purpose. Other applications might differ in their perspective and objective and deem none or other biases less or more important. As we mentioned above, since a dataset cannot maintain its status of ‘ground truth’, the responsibility of reviewing and potentially mitigating data biases falls onto researchers and practitioners.
We hope our work proves helpful in this task.

\section{Related Work}

Previous work primarily focuses on samples from minority groups and related data biases. Objectives range from mitigating such biases to improving minority coverage, i.e.~achieving better image fidelity for underrepresented data examples. Some approaches employ importance weighting where weights are derived from density ratios, either via an approximation based on the discriminator’s prediction \citep{lee2021selfdiagnosing} or via an additional probabilistic classifier \citep{grover2019bias}. Others propose an implicit maximum likelihood estimation framework to improve the overall mode coverage \citep{yu2020inclusive}. These methods either depend on additional adversarily trained models or on more specific hybrid models. 
Our approach, instead, has two major benefits over previous work. First, it is model-agnostic and thus potentially applicable to a wide range of network architectures and training schemes. Second, it only adds an offline pre-computation step prior to conventional training procedures and during training solely intervenes at the data sampling stage.

Authors of previous work further argue for increased diversity, but do not evaluate on explicit measures of diversity. Results are reported on the standard metrics \ac{IS}, as well as \ac{FID} and \ac{PR} which rely on the training dataset for reference. Consequently, they can only estimate sample fidelity and mode coverage as present in the data. 
We, instead, evaluate on measures designed to objectively quantify diversity.

Most importantly, while we argue for an adaptation of modelling techniques to allow for \emph{mode balancing} to achieve higher output diversity, all related works operate under the \emph{mode coverage} paradigm. In fact, \citet{lee2021selfdiagnosing} include a discriminator rejection sampling step \citep{azadi2018discriminator} after training to undo the bias introduced by their importance sampling scheme.

\section{Conclusions}

We introduced a method to derive a weight vector over the examples in a training dataset, which indicate their individual contribution to the dataset’s overall diversity. \emph{Diversity weights} allow to train a generative model with importance sampling such that the model’s output diversity increases.

Our work is motivated by potential benefits for computational creativity applications which aim to produce a wide range of diverse output for further design iterations, ranging from artistic over constrained to scientific creativity.
We also highlight a connection to issues of data bias in generative machine learning, in particular data imbalances and the under-representation of minority features. The impracticality of easily mitigating data imbalances in an unsupervised setting further motivates our work.

In a proof-of-concept study, we demonstrated that our method increases model output diversity when compared to a standard \ac{GAN}.
The results highlight a trade-off between artefact typicality, i.e.~the extent to which an artefact is a typical training, and diversity. Our method provides control over this trade-off via a loss balance hyperparameter.

\section{Future Work}

We plan to build on the present work in several ways.
First, by refining our method, in particular the training procedure, to improve overall sample fidelity. For this, a thorough analysis and systematic comparison to related work is needed. 
The loss balance hyperparameter could further be tuned automatically by including it as a learnable parameter in the optimisation procedure. Apart from our gradient descent approach, there might be alternative exact or approximate methods for the diversity weight optimisation, e.g.~constraint optimisation or analytical solutions.

Second, we plan to extend experimentation to other generative models and on bigger and more complex datasets to demonstrate the scalability of our approach. Since our method is architecture-agnostic, there remain many opportunities for future work to understand the effect and potential benefits of our method in other modelling techniques.
As \ac{GAN} training is notoriously unstable and requires careful tuning, other modelling techniques might prove more appropriate.
Results on datasets representing humans are needed to demonstrate the capability of our method to mitigate issues of \ac{DEI} resulting from data imbalances.

Moreover, empirical studies will be necessary to investigate how the shift from mode coverage to mode balancing can support diversity in a large range of \ac{CC} applications.

\section{Acknowledgements}
Sebastian Berns is funded by the EPSRC Centre for Doctoral Training in Intelligent Games \& Games Intelligence (IGGI) [EP/S022325/1]. 
Experiments were performed on the Queen Mary University of London Apocrita HPC facility, supported by QMUL Research-IT~\citep{king_thomas_2017_438045}.

\newpage
\section{Appendix}

The tables below outline the experiments’ training hyperparameters and network architectures, which do not include any pooling, batchnorm or dropout layers.
He initialisation (Kaiming uniform) is used for convolutional layers (conventional and upsampling) and Glorot initialisation (Xavier uniform) for fully connected (FC) layers.

\begin{table}[!ht]

\centering
\caption{Architecture of \acs{WGAN-GP} generator network. Upsampling convolutional layers (ConvTranspose) have kernel size 4~$\times$~4, stride~2, padding~1, dilation~1.}

\label{table:wgan-generator}

\begin{tabular}{c|cc}
\toprule
\multicolumn{3}{c}{\textbf{WGAN Generator}}\\
\midrule

\textbf{Layer} &
\textbf{Output} & 
\textbf{Activation} \\
\midrule

Input $z$ &
64 &
\\
 
Linear (FC) & 
2,048 & 
ReLU \\
 
Reshape & 
4 $\times$ 4 $\times$ 128 &
\\
 
ConvTranspose & 
8 $\times$ 8 $\times$ 64 & 
ReLU \\
 
Cut & 
7 $\times$ 7 $\times$ 64 & 
\\
 
ConvTranspose & 
14 $\times$ 14 $\times$ 32 & 
ReLU \\
 
ConvTranspose & 
28 $\times$ 28 $\times$ 1 & 
Sigmoid \\

\bottomrule
\end{tabular}

\end{table}

\begin{table}[!ht]

\centering
\caption{Architecture of \acs{WGAN-GP} critic network. Convolutional layers have kernel size 5 $\times$ 5, stride~2, padding~2.}

\label{table:wgan-critic}

\begin{tabular}{c|cc}
\toprule
\multicolumn{3}{c}{\textbf{WGAN Critic}}\\
\midrule

\textbf{Layer} &
\textbf{Output} & 
\textbf{Activation} \\
\midrule

Input &
28 $\times$ 28 $\times$ 1 & 
\\
 
Conv & 
14 $\times$ 14 $\times$ 32 & 
LeakyReLU(0.2) \\
 
Conv & 
7 $\times$ 7 $\times$ 64 & 
LeakyReLU(0.2) \\
 
Conv & 
4 $\times$ 4 $\times$ 128 & 
LeakyReLU(0.2) \\
 
Reshape & 
2,048 & 
\\
 
Linear (FC) & 
1 & 
\\

\bottomrule
\end{tabular}

\end{table}

\begin{table}[!ht]

\centering
\caption{Training hyperparameters}

\label{table:hyperparameters}

\begin{tabular}{ll}
\toprule
\textbf{Hyperparameter} &
\textbf{Value} \\

\midrule
Num steps &
6,000 \\
Num critic steps &
5 \\

\midrule
Batch size &
6,000 \\

\midrule
GP weight &
10.0 \\

\midrule
LR generator &
0.0001 \\
LR critic &
0.0001 \\

\midrule
Adam $\beta_1$ &
0.5 \\
Adam $\beta_2$ &
0.9 \\

\bottomrule
\end{tabular}

\end{table}

\newpage

\bibliographystyle{iccc-eprint}
\bibliography{references}

\begin{acronym}
    \acro{CC}{computational creativity}
    \acro{DL}{deep learning}
    \acro{GM}{generative model}
    \acro{ML}{machine learning}
    
    \acro{GAN}{generative adversarial network}
    \acro{VAE}{variational auto-encoder}
    
    \acro{CLIP}{contrastive language-image pre-training}
    \acro{DiaGAN}{Self-Diagnosing GAN}
    \acro{InclGAN}{Inclusive GAN}
    \acro{SAGAN}{Self-Attention Generative Adversarial Network}
    \acro{SNGAN}{Spectral Normalisation GAN}
    \acro{WGAN-GP}{Wasserstein GAN with gradient penalty}
    \acroplural{WGAN-GP}{Wasserstein GANs with gradient penalty}

    \acro{LIGM}{large image generation model}
    
    \acro{FID}{Fréchet Inception Distance}
    \acro{wFID}{Weighted Fréchet Inception Distance}
    \acro{IS}{Inception Score}
    \acro{PD}{Pure Diversity}
    \acro{PR}{Precision–Recall}
    \acro{VS}{Vendi Score}

    \acro{DivW}{diversity weights}

    \acro{DEI}{diversity, equity, and inclusion}
\end{acronym}

\end{document}